\documentclass[10pt,twocolumn,letterpaper]{article}

\usepackage[pagenumbers]{cvpr} 

\usepackage{graphicx}
\usepackage{amsmath}
\usepackage{amssymb}
\usepackage{booktabs}
\usepackage{color}
\usepackage{bbding}
\usepackage{amsmath}
\usepackage{dsfont}
\usepackage{multirow}
\usepackage{booktabs}
\usepackage{algorithm}
\usepackage{algorithmic}
\usepackage{multirow}
\usepackage[table]{xcolor}

\usepackage[pagebackref,breaklinks,colorlinks]{hyperref}

\usepackage[capitalize]{cleveref}
\crefname{section}{Sec.}{Secs.}
\Crefname{section}{Section}{Sections}
\Crefname{table}{Table}{Tables}
\crefname{table}{Tab.}{Tabs.}

\def\cvprPaperID{204} 
\def\confName{CVPR}
\def\confYear{2023}

\newtheorem{definition}{Definition}

\newtheorem{theorem}{Theorem}
\newtheorem{proposition}{Proposition}
\begin{document}

\title{EnfoMax: Domain Entropy and Mutual Information Maximization for Domain Generalized Face Anti-spoofing}

\author{Tianyi Zheng\\
Shanghai Jiao Tong University \\
{\tt\small tyzheng@sjtu.edu.cn}}
\maketitle
\begin{abstract}
  The face anti-spoofing (FAS) method performs well under intra-domain setups. However, its cross-domain performance is unsatisfactory. As a result, the domain generalization (DG) method has gained more attention in FAS. Existing methods treat FAS as a simple binary classification task and propose a heuristic training objective to learn domain-invariant features.  However, there is no theoretical explanation of what a domain-invariant feature is. Additionally, the lack of theoretical support makes domain generalization techniques such as adversarial training lack training stability. To address these issues, this paper proposes the EnfoMax framework, which uses information theory to analyze cross-domain FAS tasks. This framework provides theoretical guarantees and optimization objectives for domain-generalized FAS tasks. EnfoMax maximizes the domain entropy and mutual information of live samples in source domains without using adversarial learning. Experimental results demonstrate that our approach performs well on extensive public datasets and outperforms state-of-the-art methods.
\end{abstract}
\maketitle
\section{Introduction}
\label{sec:intro}

Face recognition (FR) techniques offer a simple yet convenient way for identity authentication applications, such as mobile access control and electronic payments. Though face biometric systems are widely used, with the emergence of various presentation attacks, critical concerns about security risks on face recognition systems are increasing. An unprotected face recognition system might be fooled by merely presenting artifacts like a photograph or video in front of the camera. Therefore, strengthening the face recognition system from various presentation attacks promotes the techniques of face anti-spoofing (FAS).

\begin{figure}[t]
\centering
\includegraphics[width=0.45\textwidth]{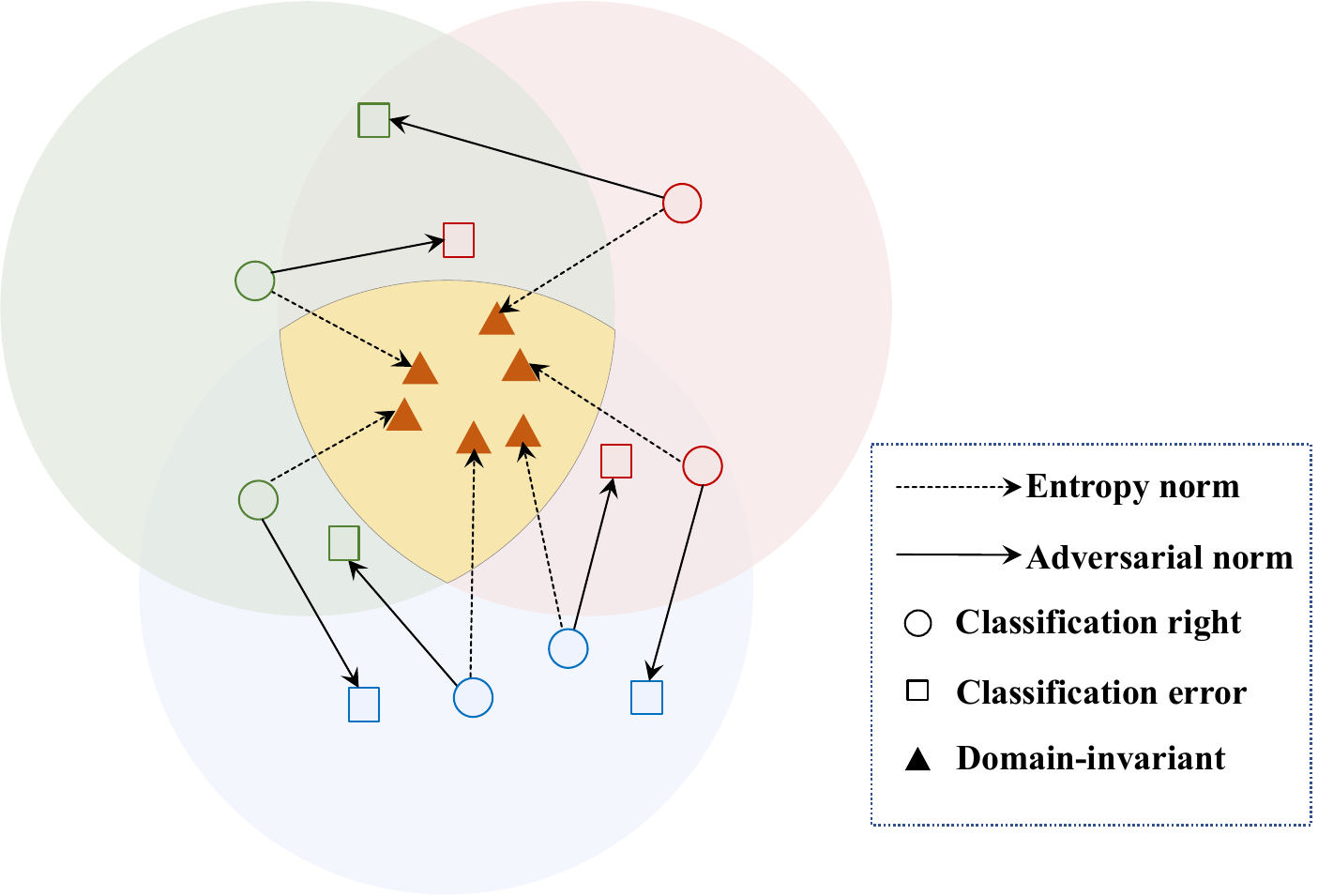} 
\caption{Comparison of adversarial learning and entropy norm methods to remove domain-special information. Different color in the figure means different domain. The adversarial norm method aims to let the domain classifier misclassify, and the domain entropy nrom method makes the domain classifier unable to distinguish the domain to which the sample belongs. }
\label{fig:entropy}
\vspace{-10pt}
\end{figure}

A series of face anti-spoofing methods have been proposed as an essential research topic, from hand-craft feature-based to deep representation-based methods. The previous FAS methods achieved promising performance. Despite its success, most current methods are heuristic and need more relevant theoretical analysis. All previous FAS methods cast the FAS task as a binary classification problem and heuristic design optimization objectives. To give a theoretical basis for the optimization objective design, we use a mutual information-based method to analyze the importance of the optimization objective design in FAS.
Meanwhile, current FAS methods may suffer dramatic degradation when encountering unseen domains. To alleviate this issue, researchers have proposed various approaches ~\cite{DBLP:conf/cvpr/Wang0SC20,DBLP:conf/aaai/QinZZWYFZSL20,DBLP:conf/aaai/ShaoLY20,Wang_2022_CVPRSSAN,liu2021dual,jia2020single,chen2021generalizable} to improve the generalizability of FAS in domain generalized settings. In the domain generalized face anti-spoofing tasks, the target domain data, and labels are unavailable in the training stages. Only source domains are available for us to find the common features of the live samples. Most previous work aims to aggregate the samples from different domains together~\cite{jia2020single,DBLP:conf/mm/Du0Z0022,DBLP:conf/wacv/LiaoCLYHC23}. Therefore, domain generalized FAS's core problem is improving the generalizability of features about live samples across different domains.

The currently available methods for addressing the generalized face anti-spoofing (FAS) problem can be broadly categorized into three types: meta-learning ~\cite{DBLP:conf/aaai/ShaoLY20,chen2021generalizable,DBLP:conf/aaai/QinZZWYFZSL20}, data augmentation~\cite{DBLP:journals/corr/abs-2106-14162,Wang_2022_CVPRSSAN} and adversarial training approaches~\cite{shao2019multi,jia2020single}. The Meta-learning based methods aim to simulate the domain shift by choosing one of the source domains as the meta-test domain. The data augmentation technique designs various augmentation methods to produce more training data and identify the shared characteristics of live samples. Adversarial training utilizes the gradient reversal layer (GRL) to increase the prediction loss of the domain discriminator, leading to the misclassification of the feature's domain.

As the domain generalized FAS task falls under the category of special domain generalization (DG) tasks, a common approach is to acquire the domain-invariant representation of samples from various domains~\cite{DBLP:conf/ijcai/0001LLOQ21}. However, despite numerous techniques designed to learn domain-invariant features of live samples in the source domains, the definition of such features remains unclear. Various methods employ different approaches to distinguish between domain-special and domain-invariant information. For instance, the SSAN~\cite{Wang_2022_CVPRSSAN} method utilizes style features as domain-special information, whereas the DiVT~\cite{DBLP:conf/wacv/LiaoCLYHC23} method considers the norm of the feature embedding of live samples across all domains as the domain-invariant information. Due to the lack of a precise definition of domain-invariant features of live samples, there exist significant differences among the various methods. 

In addition, another issue limits the generalization ability of the adversarial training methods.
Even though some previous work~\cite{Wang_2022_CVPRSSAN,jia2020single} uses adversarial training methods to make the domain discriminator predict the error domain label of the feature and achieve good performance on domain generalized FAS tasks. We argue that using the adversarial training method is insufficient for domain generalized FAS tasks. Since the previous work like ~\cite{Wang_2022_CVPRSSAN,jia2020single} uses the gradient reversal layer to make the loss function of the domain discriminator larger, the domain discriminator can only classify it from the wrong domain but cannot remove domain-special information. As depicted in Figure \ref{fig:entropy}, a sample from domain A, whether it is predicted as domain B or domain C, will both increase the prediction loss of the domain discriminator, which may lead to the introduction of domain-special information from other domains when removing their domain-special information using adversarial training. Therefore, we introduce domain entropy to measure the domain-related information contained in the feature. 

In this work, we use information theory to analyze domain generalized face anti-spoofing tasks. Instead of dividing the feature into domain-special and domain-invariant by intuition, we use mutual information and domain entropy to divide the feature. In our analysis, the domain-invariant feature should contain sufficient task information and not be predicted for its domain. So we use mutual information to measure the task-related information contained in features. Meanwhile,  we use domain entropy to measure the domain-special information in the feature. We argue that when the output of the domain classifier has the largest domain entropy, the domain-special information in the features is removed, and the features are then domain-invariant features. Our analysis provides theoretical guarantees for optimization objectives designed for domain generalized FAS based on the lower bound of the mutual information. Since our method has the guarantee of the theoretical lower bound, our method has better stability compared to other methods.
Meanwhile, to overcome the deficiency of adversarial training, We propose a constraint method based on domain entropy, which can ensure that our features have the largest entropy for domain-special features without adversarial training. Our new learning framework for domain generalized FAS tasks aims to maximize the domain entropy and mutual information of the live samples in the source domains, which called EnfoMax.

EnfoMax learning framework presents an information-theoretic approach to analyzing domain generalized FAS tasks. Compared to previous work, our method uses mutual information to measure the degree of correlation between features and labels, which provides a theoretical basis for the optimization objective design. Meanwhile, EnfoMax
uses domain entropy to measure the domain information contained in the feature. Therefore, we can use the corresponding metrics to analyze the relationship between features, task labels, and domain labels. However, another question is how to estimate mutual information between high-dimensional variables. Since the probability distribution of high-dimensional variables is usually unknown to us, fortunately, we can design the proxy task based on the variational lower bound on the mutual information to estimate the mutual information between the features and labels~\cite{DBLP:journals/corr/abs-1808-06670,DBLP:conf/icml/BelghaziBROBHC18,DBLP:journals/corr/abs-1807-03748,DBLP:conf/icml/PooleOOAT19}. Our main contributions are:
\begin{itemize}
    \item  We use mutual information and domain entropy to analyze the domain generalized FAS tasks, which provides a theoretical basis for the optimization objective design for the domain generalized FAS task.
    \item We derive a lower bound on mutual information between the feature and labels in the unseen target domain. This inspired us to design a novel learning framework called EnfoMax for domain generalized FAS tasks. Meanwhile, our EnfoMax framework overcomes the shortcomings of adversarial learning and has better training stability. 
    \item We conduct experiments on widely-used domain generalization FAS benchmarks and achieve state-of-the-art performance, which illustrates the effectiveness of our proposed EnfoMax framework. 
\end{itemize}

\section{Related Work}

\subsection{Face anti-spoofing (FAS).}
Face anti-spoofing has an essential means to protect the security of the face detection system, which can effectively improve the system's safety. The traditional features used are often hand-crafted features such as LBP~\cite{freitas2012lbp}, HOG~\cite{komulainen2013context} and SIFT~\cite{patel2016secure}. Recently, CNN-based methods~\cite{DBLP:conf/cvpr/LiuJ018,DBLP:conf/cvpr/YuZWQ0LZZ20,DBLP:conf/cvpr/WangYZZQZZL20,DBLP:conf/eccv/ZhangYLYYSL20} has been widely used in face anti-spoofing tasks. However, some recent studies~\cite{wang2022transfas, Wang_2022_CVPRSSAN} find that CNN-based methods exhibit a limited capacity to capture both local and global interdependencies within images, which are crucial face anti-spoofing. To overcome this problem, George et al.~\cite{DBLP:conf/icb/GeorgeM21} first introduce a pure vision transformer to extract features from all patches for zero-shot FAS. ViTAF method~\cite{DBLP:journals/corr/abs-2203-12175} uses transformers as the backbone and adds the adaptive module to adapt to different FAS scenarios. MTSS ~\cite{DBLP:conf/bmvc/HuangHCKLS21} method adopts vision transformer as the
teacher model to train a smaller student CNN. However, the success of these methods relies heavily on additional datasets for pre-training due to its weak inductive bias~\cite{DBLP:journals/tifs/WangWZYL22}. This paper proposes a new method to train the vision transformer for cross-domain tasks without additional datasets.

\subsection{Domain Generalization (DG)}
Domain generalization for face anti-spoofing aims to learn a model from multiple source datasets; this model can perform well in the target domain dataset. In previous work, some methods~\cite{shao2019multi,jia2020single} use adversarial training to learn a shared feature space for multiple source domains. HIRs~\cite{DBLP:conf/icpr/WangLG20} minimize the Kullback-Leibler (KL) divergence between the conditional distributions of each class. Some meta-learning formulations~\cite{DBLP:conf/aaai/ShaoLY20,chen2021generalizable,DBLP:conf/aaai/QinZZWYFZSL20} methods are exploited to simulate the domain shift at training time to learn domain-invariant representations. Meanwhile, some data augmentations~\cite{DBLP:journals/corr/abs-2106-14162,Wang_2022_CVPRSSAN} can avoid overfitting and improve the generalization of the model. Unlike the general DG methods, domain generalized FAS tasks focus on the common feature of the live samples. Therefore, an energy-based method~\cite{DBLP:conf/mm/Du0Z0022} tries to model the distribution of real samples. However, such an estimate is imprecise because of the limited number of real samples. There are also many adversarial learning-based methods~\cite{Wang_2022_CVPRSSAN,jia2020single} that try to learn the domain-invariant feature of live samples. However, what is the domain-invariant feature is very different in these methods.
This paper proposes a domain entropy maximization way to keep the domain-invariant features of live samples instead of adversarial learning.

\subsection{Mutual Information Estimation.}
Mutual information measures the relationship between different random variables~\cite{paninski2003estimation}. However, since the probability distributions are always unknown, mutual information has been difficult to compute. Recently, some neural network-based methods~\cite{DBLP:journals/corr/abs-1808-06670,DBLP:conf/icml/BelghaziBROBHC18,DBLP:journals/corr/abs-1807-03748,DBLP:conf/icml/PooleOOAT19} have been proposed to estimate the mutual information between different random variables. These methods always optimize the lower or upper bound of the mutual information between high-dimensional random variables. MINE~\cite{DBLP:conf/icml/BelghaziBROBHC18} directly uses a neural network to complete lower bound of the dual representations of the KL-divergence~\cite{e5879cd3d84b462abf51f06791e5ba28}, The method~\cite{DBLP:journals/corr/abs-1807-03748} propose the lower bound of mutual information based on infoNCE loss. Meanwhile, some work~\cite{DBLP:conf/nips/Tian0PKSI20,tsai2020self} also uses contrastive learning-based methods to bind the mutual information. In the domain generalize tasks, MIRO~\cite{DBLP:conf/eccv/ChaLPC22} re-formulate the DG objective using mutual information with the oracle model, Another method~\cite{DBLP:conf/cvpr/RagonesiVCM21} minimizes the mutual information between the learned representation and specific data attributes. In addition, the adversarial ~\cite{DBLP:conf/nips/MoyerGBGS18} approach can also be derived as a proxy task to minimize the mutual information. Since many proxy tasks bind the mutual information, we leverage different proxy tasks to estimate the mutual information in the domain generalized FAS tasks.

\section{Proposed Method}
\subsection{Problem Formulation}
The following is a formal definition of the domain generalized FAS task: Given a collection of domains $D$, only a subset is available for training. We aim to learn an encoder $G_{\theta}$, which can effectively generalize to unseen domains based on the information from the seen domains. We denote the seen domains as $D^{s}$ and the unseen domains as $D^{u}$. Each domain encompasses two data types in the FAS task: live and spoof samples, represented as $Y_{live}$ and $Y_{spoof}$, respectively. Furthermore, every source dataset possesses a domain label, denoted as $Z_{j}$. We use $D^{s}(X, Y, Z)$ to represent our source domain dataset. Concurrently, we utilize $T$ as the representative feature of the $X$. In the domain generalized FAS task setting, we do not have any information about the target domain during the training stage.

\subsection{Information Theoretic Analysis}
This subsection gives an information theoretic analysis of the domain generalized FAS task. Our goal is to minimize the classification error of the model on the unknown domain $D^{u}$. Under the information-theoretic view, this optimization problem can be transformed into the following problem:
\begin{equation}
    \label{final_goal}
    max \ I^{u}(Y_{live};T)
\end{equation}
The $I^{u}(Y_{live};T)$ is the mutual information between the labels of live faces and the feature extracted by the encoder $G_{\theta}$. Solving (\ref{final_goal}) is the proxy task~\cite{DBLP:conf/ijcai/0001LLOQ21} to minimize the classification error.  However, we know nothing about the label and data information of the unseen domain $D^{u}$. Therefore, most existing general domain generalization methods aim to find the domain-invariant features of each class in the source domains~\cite{DBLP:conf/nips/NguyenTGB21}. However, in the FAS cross-domain task, we only want to find the common features of the live samples. This is the main difference between domain generalized FAS and other domain generalization tasks. Then we give some useful definitions and propositions of our method.
\begin{definition}[Domain gap between live samples]
When an encoder $G_{\theta}$ trained on the seen domain is given, the live samples domain gap between the unseen domain and the seen domain is defined as:
\label{def1}
\begin{equation}
\Gamma (G_{\theta}|Y_{live})=D(p^{u}(T|Y_{live})||p^{s}(T|Y_{live}))
\end{equation}
\end{definition}
The ideal situation is no domain gap between live faces when the $G_{\theta}$ is trained. It means that the feature extracted by the encoder $G_{\theta}$ are aligned without bias. In that case, the encoder $G_{\theta}$ learns the domain-invariant features of the source domain. Since the encoder $G_{\theta}$ is trained on the seen domains $D^{s}$, we have the proposition (\ref{prop1}).
\begin{proposition}
An encoder $G_{\theta}$ trained on the seen domain, the represent feature $T$, and label $Y_{live}$ have more mutual information in seen domain than the unseen domain. 
\label{prop1}
\begin{equation}
    I^{s}(Y;T)>I^{u}(Y;T))
\end{equation}
\end{proposition}




\begin{definition}
Let $\Psi\left(\delta\mid Y_{live}\right)$ denote the supremum discrepancy of mutual information between live samples in seen domain and unseen domain when $\Gamma (G_{\theta}|Y_{live}) \geq \delta$. i.e.,
\label{def3}
\begin{equation}
  \begin{aligned}
\Psi\left(\delta \mid Y_{live}\right)= & \sup \left|I^s\left(Y ; T\right)-I^u\left(Y_{live} ; T\right)\right| \\
\end{aligned}
\end{equation}
\end{definition}
When the domain gap $\delta$ between live samples becomes larger, the supremum discrepancy of mutual information between live samples in seen domain and unseen domain also becomes larger, indicating that $\Psi$ is a non-decreasing function of $\delta$. In addition, this item also measures the domain-special features contained in $T$. 

Based on the definitions and proposition, we get our main result in Theorem \ref{thm:low}.
\begin{theorem}
For seen domain $D^{s}$, unseen domain $D^{u}$ and the encoder $G_{\theta}$, the following inequalities hold. 
\label{thm:low}
\begin{equation}
 \label{opobj}
  \begin{aligned}
I^{u}(T;Y_{live})\geq & I^{s}(T;Y) -\Psi(\Gamma (G_{\theta}|Y_{live}))
\end{aligned}
\end{equation}
\end{theorem}
The proofs are provided in the \textbf{appendix} {\color{red} sec.1}. By the Thm (\ref{thm:low}), we have a possible way to maximize the lower bound of the mutual information proposed in (\ref{final_goal}). The first item $I^{s}(T; Y)$ measures the task-related information in the feature $T$, which we want to maximize. As for the second item $\Psi(\Gamma (G_{\theta}|Y_{live}))$, this item measures the discrepancy between the source and target domains, which can be used to measure the domain-special features in $T$. Therefore, our purpose is to minimize it. Since we know nothing about the target domain, we can maximize the average live sample domain dispersibility as the proxy task. Optimizing it can remove the domain-special information contained in the feature $T$. Meanwhile, we need to add label information to ensure that we are constraining live samples. Therefore, we add a regular item that maximizes the mutual information between $T$ and $Y$. 

We first prove that optimizing the reconstruction task equals maximizing the mutual information between the $T$ and $Y$~\cite{barber2004algorithm}. We use $\Tilde{X}$ to denote the reconstructed face image. Since $\Tilde{X}$ and $X$ have the label $Y$, we can use $\Tilde{X}$ to indicate the label $Y$.

Based on the definition of the mutual information~\cite{DBLP:books/daglib/0016881}, the mutual information between the $X$ and $\Tilde{X}$ is given in Equation (\ref{eq:mi}):
\begin{equation}\label{eq:mi}
\begin{split}
I\left(\Tilde{X};G_{\theta}(X)\right)=&H\left(\Tilde{X}\right)-H\left(\Tilde{X} \mid G_{\theta}(X)\right)\\=&H(G_{\theta}(X))-H\left(G_{\theta}(X) \mid \Tilde{X}\right).
\end{split}
\end{equation}
By the definition of conditional entropy, we have:
$$
H\left(\Tilde{X} \mid G_{\theta}(X)\right)=\mathbb{E}_{P_{\Tilde{X}, G_{\theta}(X)}}\left[-\log P\left(\Tilde{X} \mid G_{\theta}(X)\right)\right].
$$
But in practice, it's difficult to get the distribution of $P\left(\Tilde{X} \mid G_{\theta}\left(X\right)\right)$ directly. The most common way to approximate this distribution is using another distribution $Q(x)$ instead of it and maximizing the lower bound of KL divergence between them~\cite{agakov2004algorithm}:
\begin{equation}
\begin{aligned}
\label{low1}
&I\left(G_{\theta}\left(X\right);\Tilde{X}\right) \\
&=H\left(\Tilde{X}\right)-H\left(\Tilde{X} \mid G_{\theta}\left(X\right)\right) \\
&=H\left(\Tilde{X}\right)+\mathbb{E}_{P_{\Tilde{X}, G_{0}\left(X\right)}}\left[\log P\left(\Tilde{X} \mid G_{\theta}\left(X\right)\right)\right] \\
&=H\left(\Tilde{X}\right)+\mathbb{E}_{P_{\Tilde{X}, G_{\theta}\left(X\right)}}\left[\log Q\left(\Tilde{X} \mid G_{\theta}\left(X\right)\right)\right] \\
& \;+\underbrace{D_{K L}\left(P\left(\Tilde{X} \mid G_{\theta}\left(X\right)\right) \| Q\left(\Tilde{X} \mid G_{\theta}\left(X\right)\right)\right)}_{\geq 0} \\
& \geq \mathbb{E}_{P_{\Tilde{X}, G_{0}\left(X\right)}}\left[\log Q\left(\Tilde{X} \mid G_{\theta}\left(X\right)\right)\right].
\end{aligned}
\end{equation}

The distribution $Q(x)$ can be chosen arbitrarily. So we can use Gaussian distribution with $\sigma I$  diagonal matrix as $Q(x)$ ~\cite{ly2022student} i.e. $Q(x)\sim\mathcal{N}\left(\Tilde{X} \mid G_{\theta}\left(X\right), \sigma I\right)$. Therefore, the maximize problem can be converted to the minimal problem in Equation (\ref{eq:min}):
\begin{equation}\label{eq:min}
\min  \mathbb{E}_{P_{\Tilde{X}, X}}[\left\| D_{\theta}\left(G_{\theta}(X) \right)-\Tilde{X}\right\|_{2}^{2}].
\end{equation}

In the reconstruction task, we want to minimize the MSE loss of each pixel. 
Next, we prove that $T$ can be used as the representation of the original image. Based on the assumption proposed by ~\cite{DBLP:conf/colt/SridharanK08}, we know that the reconstructed image and input image are both redundant for the task-relevant information, i.e., there exists an $\epsilon$ s.t. $I(\Tilde{X};T|D_{\theta}(G_{\theta}(X)) \leq \epsilon$.

\begin{theorem}
\label{thmself}
The learned aggregate token contains all the task-relevant information~\cite{DBLP:conf/iclr/Tsai0SM21} in the input $X$ with a potential loss $\epsilon$.
\begin{equation}
I(\Tilde{X};G_{\theta}(X)) -  \epsilon \leq  I(T;G_{\theta}(X)) \leq I(\Tilde{X};G_{\theta}(X)).
\end{equation}
\end{theorem}

The proofs are provided in the \textbf{appendix} {\color{red} sec.2}. By Thm (\ref{thmself}). we can get the result that $T$ can be used as the original image's task-relevant representation. The ideal situation is that $T$ is the sufficient statistic for estimating the $\Tilde{X}$. In that case, $G_{\theta}(X)$, $T$ and $\Tilde{X}$ form a Markov chain $\Tilde{X} \leftrightarrow T \leftrightarrow G_{\theta}(X)$ and $T$ can represent $\Tilde{X}$ without any information loss. Therefore, we can use $T$ instead of $G_{\theta}(X)$ as the representative features.

Since we only have the source domain data in our training stage, we want to align the feature of live samples of the different domains. In other words, we want to remove the domain-special features in $T$. By information theory, entropy effectively measures the degree of chaos in a system. Therefore, we use domain entropy to measure the domain-related information contained in the feature given by the encoder $G_{\theta}$. When the $T$ is used to predict the domain label of input live samples, we want the predicted result of the domain to have the maximum domain entropy.          
\begin{figure*}[ht]
\centering
\includegraphics[width=0.9\textwidth]{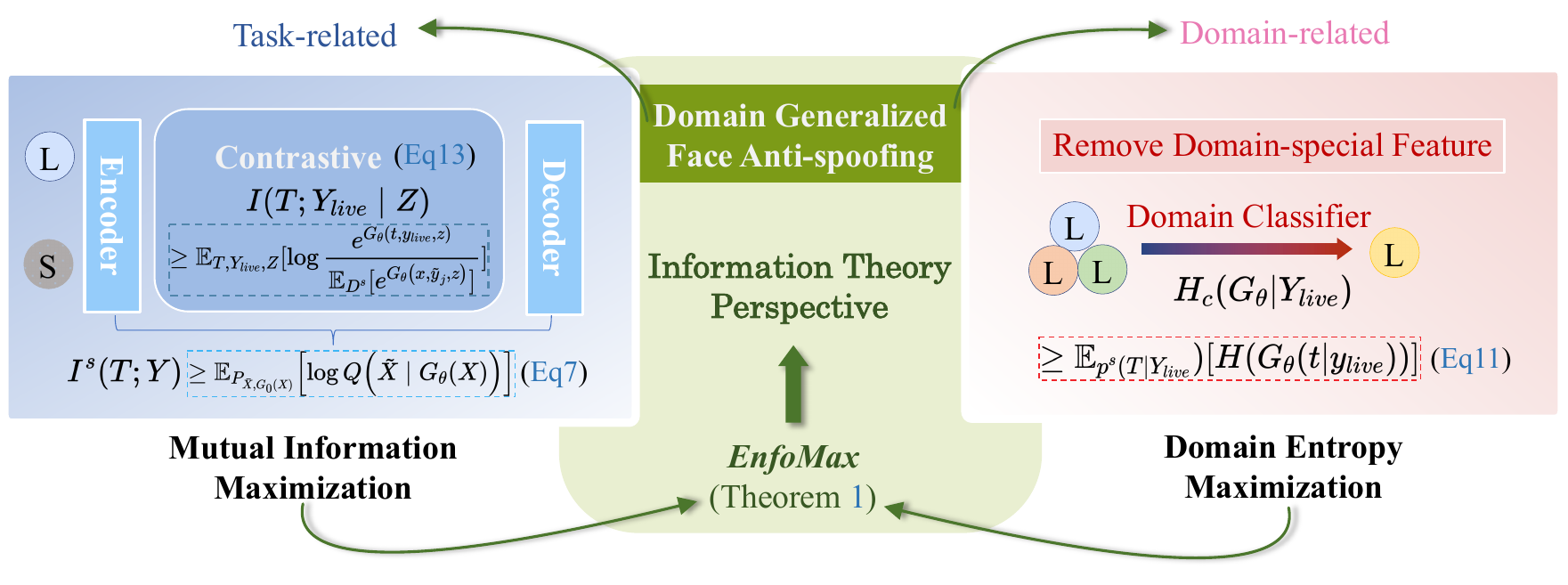} 
\caption{Overview of proposed the EnfoMax framework for domain generalized FAS. Our framework uses mutual information to measure the task-related feature and uses domain entropy to measure the domain-related feature. To optimize the lower bound of each term, we design the corresponding module.}
\label{fig:method}
\end{figure*}
\begin{definition}
$D^{s}(X, Y_{live}, Z)$ are the source datasets of FAS tasks that have live labels. We use averaged domain entropy to measure the domain-related information given by the encoder $G_{\theta}$ in each source domain:
\label{def4}
\begin{equation}
\begin{split}
H_{c}(G_{\theta}|Y_{live})  \triangleq  \frac{1}{|Z|} & \sum_{j=1}^{|Z|} \mathbb{E}_{p^{s}(T|Y_{live}}) \cdot \\ & H\left(\frac{\sum_{i=1}^{|Z_{j}|} G_{\theta}(t|y_{live}),z_{j})}{|Z_{j}|}\right)
\end{split}
\end{equation}
\end{definition}
In summary, averaged domain entropy depicts the average degree of dispersion of the average prediction of samples in each domain. Maximizing it equals removing the domain-related information in each domain. Meanwhile, since entropy is a concave function, we have the proposition (\ref{prop2}).
\begin{proposition}
The averaged domain entropy of live samples is always greater than or equal to the average live sample domain entropy, i.e.,
\begin{equation}
\label{low2}
H_{c}(G_{\theta}|Y_{live}) \geq \mathbb{E}_{p^{s}(T|Y_{live}}) [H(G_{\theta}(t|y_{live}))]
\end{equation}
\label{prop2}

\end{proposition}
This proposition can be proved by Jensen's inequality. Equality holds if and only if the feature of each sample extracted by the encoder $G_{\theta}$ is predicted to have the same probability in different source domains. Therefore, we can optimize the lower bound of averaged domain entropy of live samples $H_{c}(G_{\theta}|Y_{live})$ by maximizing the average live sample domain entropy. Therefore, for the second item in Thm (\ref{thm:low}), we have the following objective to optimizer:
\begin{equation}
    \label{domain}
    max\ \mathbb{E}_{p^{s}(T|Y_{live}}) H_{d}(G_{\theta}|Y_{live}) + \lambda_{2} I(Y_{live};T|Z)  
\end{equation}
This objective contains two items. The first item is the average live sample domain entropy, which takes the maximum value when predicting each live sample uniformly distributed in each domain. The domain-special features are removed from the live sample when the maximum is achieved. The second item is the mutual information between $T$ and $Y$ in each source domain; we can optimize the lower bound of it via the contrastive method.
\begin{proposition}[Lower Bound for $ I(T;Y_{live}|Z)$]
For the source dataset $D^{s}(X,Y,Z)$ of FAS, we have the lower bound of $I(T;Y_{live}|Z)$ via contrastive estimation~\cite{DBLP:conf/icml/PooleOOAT19}
\begin{equation}
\label{domain}
I(T;Y_{live} \mid Z) \geq \mathbb{E}_{T, Y_{live}, Z} [\log \frac{e^{G_{\theta}(t, y_{live}, z)}}{ \mathbb{E}_{D^{s}} [e^{G_{\theta}\left(x,\tilde{y}_j, z\right)}]}]
\end{equation}
$G_{\theta}$ is the encoder; The lower bound is the conditional version of the previous work~\cite{DBLP:journals/corr/abs-1807-03748}. In the domain generalized FAS task, the contrastive estimation only uses the live sample as the anchor.

\label{thm4}

\end{proposition}


\subsection{Overall Obejective}
Based on the above definition and proposition, we can write our objective, i.e., the lower bound of the Equation \ref{opobj} as:
\begin{equation}
\begin{split}
    max\ \underbrace{I^{s}(T;Y)}_{Eq.\ref{low1}} &+  \lambda_{1}\underbrace{\mathbb{E}_{p^{s}(T|Y_{live}}) H_{d}(G_{\theta}|Y_{live})}_{Eq.\ref{low2}} \\ &+ \lambda_{2} \underbrace{I(Y_{live};T|Z)}_{Eq.\ref{domain}}  
\end{split}
\end{equation}
The first item is lower bounded by the reconstruction loss in Equation (\ref{low1}), the second item is lower bounded by the average live sample domain dispersibility, and the last item is lower bound by the contrastive loss in  Equation (\ref{domain}).

\begin{table*}[]
\scalebox{0.8}{
\centering
\begin{tabular}{c|cc|cc|cc|cc|cc}
\hline
\multirow{2}{*}{Method} & \multicolumn{2}{c|}{O\&C\&I to M}                   & \multicolumn{2}{c|}{O\&M\&I to C}                   & \multicolumn{2}{c|}{O\&C\&M to I}                   & \multicolumn{2}{c|}{I\&C\&M to O}                    & \multicolumn{2}{c}{Average}                         \\ \cline{2-11} 
                        & \multicolumn{1}{c|}{HTER(\%)}      & AUC(\%)        & \multicolumn{1}{c|}{HTER(\%)}      & AUC(\%)        & \multicolumn{1}{c|}{HTER(\%)}      & AUC(\%)        & \multicolumn{1}{c|}{HTER(\%)}       & AUC(\%)        & \multicolumn{1}{c|}{HTER(\%)}      & AUC(\%)        \\ \hline
MADDG  ~(\cite{shao2019multi})                 & \multicolumn{1}{c|}{17.69}         & 88.06          & \multicolumn{1}{c|}{24.50}         & 84.51          & \multicolumn{1}{c|}{22.19}         & 84.99          & \multicolumn{1}{c|}{27.98}          & 80.02          & \multicolumn{1}{c|}{23.09}         & 84.40          \\
NAS-FAS   ~(\cite{yu2020fas})              & \multicolumn{1}{c|}{16.85}         & 90.42          & \multicolumn{1}{c|}{15.21}         & 92.64          & \multicolumn{1}{c|}{11.63}         & 96.98          & \multicolumn{1}{c|}{13.16}          & 94.18          & \multicolumn{1}{c|}{14.21}         & 93.56          \\
$D^{2}$AM     ~(\cite{chen2021generalizable}  )                    & \multicolumn{1}{c|}{12.70}         & 95.66          & \multicolumn{1}{c|}{20.98}         & 85.58          & \multicolumn{1}{c|}{15.43}         & 91.22          & \multicolumn{1}{c|}{15.27}          & 90.87          & \multicolumn{1}{c|}{16.10}         & 90.83          \\
SSDG-R  ~(\cite{jia2020single})                & \multicolumn{1}{c|}{7.38}          & 97.17          & \multicolumn{1}{c|}{10.44}         & 95.94          & \multicolumn{1}{c|}{11.71}         & 96.59          & \multicolumn{1}{c|}{15.61}          & 91.54          & \multicolumn{1}{c|}{11.29}         & 95.31          \\
ANRL       ~(\cite{liu2021adaptive})             & \multicolumn{1}{c|}{10.83}         & 96.75          & \multicolumn{1}{c|}{17.83}         & 89.26          & \multicolumn{1}{c|}{16.03}         & 91.04          & \multicolumn{1}{c|}{15.67}          & 91.90          & \multicolumn{1}{c|}{15.09}         & 92.24          \\
DRDG    ~(\cite{liu2021dual} )                 & \multicolumn{1}{c|}{12.43}         & 95.81          & \multicolumn{1}{c|}{19.05}         & 88.79          & \multicolumn{1}{c|}{15.56}         & 91.79          & \multicolumn{1}{c|}{15.63}          & 91.75          & \multicolumn{1}{c|}{15.67}         & 92.04          \\
SSAN-R   ~(\cite{Wang_2022_CVPRSSAN}   )               & \multicolumn{1}{c|}{6.67}          & 98.75          & \multicolumn{1}{c|}{10.00}         & \textbf{96.67} & \multicolumn{1}{c|}{8.88}          & 96.79          & \multicolumn{1}{c|}{13.72}          & 93.63          & \multicolumn{1}{c|}{9.82}          & 96.46          \\
PatchNet  ~(\cite{Wang_2022_CVPR}    )              & \multicolumn{1}{c|}{7.10}          & 98.46          & \multicolumn{1}{c|}{11.33}         & 94.58          & \multicolumn{1}{c|}{13.40}         & 95.67          & \multicolumn{1}{c|}{11.82}          & 95.07          & \multicolumn{1}{c|}{10.91}         & 95.95          \\ \hline
ViTAF ${^\dag}$ ~(\cite{huang2022adaptive}  )                    & \multicolumn{1}{c|}{4.75}          & 98.79          & \multicolumn{1}{c|}{15.70}         & 92.76          & \multicolumn{1}{c|}{17.68}         & 86.66          & \multicolumn{1}{c|}{16.46}          & 90.37          & \multicolumn{1}{c|}{13.65}         & 92.15          \\ 
TTN-T ~(\cite{DBLP:journals/tifs/WangWDG22}  )                    & \multicolumn{1}{c|}{11.25}          & 95.08          & \multicolumn{1}{c|}{11.30}         & 95.33          & \multicolumn{1}{c|}{15.75}         & 91.25          & \multicolumn{1}{c|}{14.44}          & 93.50          & \multicolumn{1}{c|}{13.19}         & 93.79 \\
TransFAS~(\cite{wang2022transfas})                & \multicolumn{1}{c|}{7.08}          & 96.69          & \multicolumn{1}{c|}{9.81}          & 96.13          & \multicolumn{1}{c|}{10.12}         & 95.53          & \multicolumn{1}{c|}{15.52}          & 91.10          & \multicolumn{1}{c|}{10.63}         & 94.86          \\
DiVT-T ~(\cite{DBLP:conf/wacv/LiaoCLYHC23} )                & \multicolumn{1}{c|}{7.14}          & 98.27          & \multicolumn{1}{c|}{11.89}         & 95.17          & \multicolumn{1}{c|}{11.43}         & 97.00          & \multicolumn{1}{c|}{15.42}          & 92.97          & \multicolumn{1}{c|}{11.47}         & 95.85          \\
\textbf{EnfoMax}       & \multicolumn{1}{c|}{\textbf{2.86}} & \textbf{99.10} & \multicolumn{1}{c|}{\textbf{8.67}} & 96.14          & \multicolumn{1}{c|}{\textbf{4.42}} & \textbf{98.07} & \multicolumn{1}{c|}{\textbf{10.92}} & \textbf{95.18} & \multicolumn{1}{c|}{\textbf{6.72}} & \textbf{97.29} \\ \hline
\end{tabular}
}
\caption{Comparison results between our EnfoMax method and state-of-the-art methods on cross-dataset testing. ViTAF${^\dag}$ denote the ViT-Base model pre-trained by the ImageNet dataset.}
\label{table:result}
\vspace{-15pt}
\end{table*}

\subsection{Our Approach}

Motivated by Theorem (\ref{thm:low}) and the Equation (\ref{opobj}), we propose our framework in Figure \ref{fig:method}. As shown in Figure \ref{fig:method}, our framework consists of three parts, i.e., the reconstruct module for the proxy task of Equation (\ref{low1}), the Domain classification module for the proxy task of Equation (\ref{low2}), and the contrastive module
for the proxy task of Equation (\ref{domain}).

\noindent \textbf{Reconstruct Module.}
In this module, we use the MIM method as the reconstruct module. Following the previous work~\cite{MaskedAutoencoders2021}, we divide the input image into non-overlapping patches and then project each into tokens $T_{i}$. Then we randomly choose a subset of the token sequence $\{T_{v_{i}}\}$ to keep, and the other tokens are masked and denoted as  $\{T_{m_{i}}\}$. The $\{T_{v_{i}}\}$ are fed into the encoder $G_{\theta}$ to get the latent representation of the original image. we use aggregation to represent the image's features, denoted as $T$. The decoder $D_{\theta}$ combine $\{T_{v_{i}}\}$ with $\{ T_{m_{i}}\}$ to reconstruct the original input image. We compute the MSE on masked tokens.
i.e.
\begin{equation} \label{eq:mse}
\mathcal{L}_{rec}=\frac{1}{n}  \sum_{i=1}^{n}\left\|D_{\theta}(G_{\theta}\left(T_{v}\right),T_{m})-T_{i}\right\|_{2}^{2} \mathds{1}_{mask}(i).
\end{equation}
The indicate function $\mathds{1}_{mask}(i)$ indicates whether token $T_{i}$ is masked is defined as Equation (\ref{eq:indicate}):
\begin{equation} \label{eq:indicate}
\mathds{1}_{mask}(i)=\begin{cases}
1,  &\quad i\in T_{m}  \\
0, &\quad  i \notin T_{m}
\end{cases}.
\end{equation}

\noindent \textbf{Contrastive Module.}
We use the contrastive module to maximize the lower bound of $ I(T;Y_{live}|Z)$. Unlike other contrastive learning methods that use different augmented to construct positive samples, our contrastive module aims to maximize the lower bound of $ I(T;Y_{live}|Z)$. This means that we need to choose live samples based on the label. Therefore, we use a supervised contrastive module in our framework. Specially, 
\begin{equation} \label{eq:conloss}
\begin{split}
\mathcal{L}_{con} = &-\mathbb{E}_{D}[\sum_{j=1}^{N}\mathds{1}_{i \neq j}(1-\mathds{1}_{y_{i} \neq y_{j}} )\\ &\log \frac{ \exp  \left(s_{i, j} / \tau\right)}{ \exp \left(s_{i, j} / \tau\right)+\sum_{k=1}^{N} \mathds{1}_{y_{i} \neq y_{k}} \exp \left(s_{i, k} / \tau\right)}].
\end{split}
\end{equation}
In the above Equation (\ref{eq:conloss}), $N$ is the mini-batch size; $y_{i}$ and $y_{j}$ mean the label of sample $i$  and sample $j$;  $\tau$ is the temperature parameter; $\mathds{1}_{y_{i} \neq y_{j}}$ is the indicator function of whether sample $i$ and sample $j$ have the same label; $s_{i j}$ is the cosine similarity between sample $i$ and sample $j$. Every live sample from different source domain in a mini-batch is used as an anchor once.

\noindent \textbf{Domain Classification Module.}
We propose this module to maximize the average live sample domain entropy. By the definition of entropy~\cite{DBLP:books/daglib/0016881}, maximizing the average live sample domain entropy equals minimizing its KL divergence with the uniform distribution. Therefore, we design a domain classifier after the encoder to classify the domain to which live samples belong. This can be achieved by minimizing the KL divergence between the output of the domain classifier and the uniformly distributed. i.e.
\begin{equation}
   \mathcal{L}_{d}=\mathbb{E}_{D}[D_{KL}(P_{d}(T|y_{live})||U_{d}(s|y_{live}))] 
\end{equation}
$P_{d}$ is the output of the domain classifier, and $U_{d}(s)$ is the uniform distribution of the domain label in the source domain.

 \begin{table}[]

\scalebox{0.8}{
\centering
\begin{tabular}{c|cc|cc}
\hline
\multirow{2}{*}{Method} & \multicolumn{2}{c|}{M\&I to C}                       & \multicolumn{2}{c}{M\&I to O}                        \\ \cline{2-5} 
                        & \multicolumn{1}{c|}{HTER(\%)}       & AUC(\%)        & \multicolumn{1}{c|}{HTER(\%)}       & AUC(\%)        \\ \hline
IDA ~(\cite{wen2015face}  )                    & \multicolumn{1}{c|}{45.16}          & 58.80           & \multicolumn{1}{c|}{54.52}          & 42.17          \\
MADDG    ~(\cite{shao2019multi})               & \multicolumn{1}{c|}{41.02}          & 64.33          & \multicolumn{1}{c|}{39.35}          & 65.10           \\
SSDG    ~(\cite{jia2020single}  )            & \multicolumn{1}{c|}{31.89}          & 71.29          & \multicolumn{1}{c|}{36.01}          & 66.88          \\
DR-MD-Net   ~(\cite{wang2020cross}  )            & \multicolumn{1}{c|}{31.67}          & 75.23          & \multicolumn{1}{c|}{34.02}          & 72.65          \\
ANRL  ~(\cite{liu2021adaptive} )                  & \multicolumn{1}{c|}{31.06}          & 72.12          & \multicolumn{1}{c|}{30.73}          & 74.10           \\
SSAN     ~(\cite{Wang_2022_CVPRSSAN} )             & \multicolumn{1}{c|}{30.00}             & 76.20           & \multicolumn{1}{c|}{29.44}          & 76.62          \\ \hline
\textbf{EnfoMax (Ours)}              & \multicolumn{1}{c|}{\textbf{29.89}} & \textbf{77.65} & \multicolumn{1}{c|}{\textbf{22.00}} & \textbf{83.56} \\ \hline
\end{tabular}
}
\caption{The result of the limited source domain experiments. Even though the dataset is limited, our method still gets good performance in each setting. }
\label{tab:limited}
\vspace{-20pt}
\end{table}

\subsection{Training and inference }
The above components together our framework. Since we use the MIM framework in our training stage, our training stage is contained by the pre-training and fine-tuning stages. We use source domain datasets to pre-train and fine-tune our model. More details of the experiments are in the \textbf{appendix} {\color{red} sec.3}.

\noindent \textbf{Pre-training stage} In the Pre-training stage, we combine the three components as our final loss function. Thus, the overall objective to be minimized when pre-training the model is
\begin{equation}
\mathcal{L} = \mathcal{L}_{rec} + \lambda_{1} \mathcal{L}_{con} + \lambda_{2}\mathcal{L}_{d}
\end{equation}
In the Pre-training stage, $\lambda_{1}$ is set to 0.1 and $\lambda_{2}$ is set to 1. We pre-train our model 400 epochs in each setting.

\noindent \textbf{Fine-tuning stage} In the fine-tuning stage, we utilize the trained encoder as our feature extractor and discard the decoders and domain classifiers. We feed the original face images into the encoder without masking and adopt the binary cross-entropy function as the loss function during the fine-tuning stage. 

\noindent \textbf{Inference stage.} In the inference stage, only the encoder $G_{\theta}$ is kept for testing the result. We also use the original face image without masking as input and get the inference result.

\begin{table*}[]
\scalebox{0.9}{
\centering
\begin{tabular}{ccc|cc|cc|cc|cc}
\hline
\multicolumn{3}{c|}{Module}       & \multicolumn{2}{c|}{O\&C\&I to M}                   & \multicolumn{2}{c|}{O\&M\&I to C}                   & \multicolumn{2}{c|}{O\&C\&M to I}                   & \multicolumn{2}{c}{I\&C\&M to O}                     \\ \hline
$\mathcal{L}_{rec}$         & $\mathcal{L}_{con}$       & $\mathcal{L}_{d}$         & \multicolumn{1}{c|}{HTER(\%)}      & AUC(\%)        & \multicolumn{1}{c|}{HTER(\%)}      & AUC(\%)        & \multicolumn{1}{c|}{HTER(\%)}      & AUC(\%)        & \multicolumn{1}{c|}{HTER(\%)}       & AUC(\%)        \\ \hline
\Checkmark          &           &           & \multicolumn{1}{c|}{6.19}          & 96.98          & \multicolumn{1}{c|}{16.67}         & 90.79          & \multicolumn{1}{c|}{12.67}         & 94.97          & \multicolumn{1}{c|}{13.19}          & 92.92          \\
  \Checkmark        &    \Checkmark       &           & \multicolumn{1}{c|}{5.71}          & 96.08          & \multicolumn{1}{c|}{11.78}         & 93.83          & \multicolumn{1}{c|}{7.42}          & 95.39          & \multicolumn{1}{c|}{13.14}          & 93.09          \\
     \Checkmark     &           &    \Checkmark       & \multicolumn{1}{c|}{5.71}          & 98.69          & \multicolumn{1}{c|}{9.33}          & 95.91          & \multicolumn{1}{c|}{5.83}          & 97.29          & \multicolumn{1}{c|}{12.72}          & 93.46          \\
\textbf{\Checkmark} & \textbf{\Checkmark} & \textbf{\Checkmark} & \multicolumn{1}{c|}{\textbf{2.86}} & \textbf{99.10} & \multicolumn{1}{c|}{\textbf{8.67}} & \textbf{96.14} & \multicolumn{1}{c|}{\textbf{4.42}} & \textbf{98.07} & \multicolumn{1}{c|}{\textbf{10.92}} & \textbf{95.18} \\ \hline
\end{tabular}
}
\caption{Effectiveness of each proposed component of EnfoMax. }
\label{table:ab_module}
\vspace{-16pt}
\end{table*}

\section{Experiments}
\subsection{Experimental Setups}
\textbf{Experiment Datasets.}
We evaluate proposed methods on cross-dataset testing based on four public datasets, CASIA-MFSD(C)~\cite{zhang2012face}, Replay-Attack(I)~\cite{chingovska2012effectiveness}, MSU-MFSD(M)~\cite{wen2015face} and OULU-NPU(O)~\cite{boulkenafet2017oulu}. Since various devices sample each dataset in different scenarios, there are large differences between those datasets. We treat each dataset as one domain. Experiments in such a setting can be a good evaluation of the model's generalization ability.

\noindent \textbf{Implementation Details.}
We use MTCNN~\cite{zhang2016joint} to detect faces in each dataset, then crop and resize each face image into 256 $\times$ 256 $\times$ 3. We use ViT-Tiny as our backbone, whose embedding dimension is 192, and the patch size of each image is 16 $\times$ 16. We only use random resized cropping as our data augmentation method. We use the same evaluation metric as previous work~\cite{shao2019multi}, i.e., the Half Total Error Rate (HTER) and the Area Under Curve (AUC).

\subsection{Method effectiness}
\textbf{Cross-domain experiments.} For the cross-domain experiments, we follow
the previous domain generalization setting. We use Leave-One-Out (LOO) setting to do our cross-domain experiments. In this evaluation protocol, we train our model using three source datasets and test our model on the remaining dataset. The result compared to other state-of-the-art methods are shown in Table \ref{table:result}. Our method performs best on all metrics except protocol O\&M\&I to C, where AUC ranks second. However, we have HTER improved compared to SSAN~\cite{Wang_2022_CVPRSSAN}, which performs best in the AUC of protocol O\&M\&I to C. Notably, EnfoMax achieves the best overall performance compared to other methods.
   
Table \ref{table:result} indicates that EnfoMax demonstrates strong generalization ability, consistently outperforming all ViT models across all settings. Even though some methods such as PatchNet~\cite{Wang_2022_CVPR} and DiVT~\cite{DBLP:conf/wacv/LiaoCLYHC23} use the extra label, EnfoMax still performance better, meanwhile, other ViT architecture models ViTAF~\cite{huang2022adaptive} DivT~\cite{DBLP:conf/wacv/LiaoCLYHC23} and TransFAS~\cite{wang2022transfas} use the extra dataset to pre-train their model, EnfoMax outperforms these models without any additional data.

\noindent \textbf{Limited source domain.}
This is a more challenging setting since only two source datasets are available in the training stage. We train our model on M and I and test our model on O and C. We show our results in Table \ref{tab:limited}. Even though the source data is limited, EnfoMax still performs best, which shows the good generalization of EnfoMax. In addition, the EnfoMax method has significant performance improvement on protocol M\&I to O.

\begin{table}
    \centering
	\begin{minipage}{0.40\linewidth}
	\centering
\scalebox{0.6}{
\begin{tabular}{c|cc}
\hline
\multirow{2}{*}{Method} & \multicolumn{2}{c}{I\&C\&M to O}        \\ \cline{2-3} 
                        & \multicolumn{1}{c|}{HTER(\%)} & AUC(\%) \\ \hline
None                    & \multicolumn{1}{c|}{12.72}    & 93.46   \\
SimCLR~\cite{DBLP:conf/icml/ChenK0H20}                  & \multicolumn{1}{c|}{12.36}    & 93.75   \\ 
SimSiam~\cite{DBLP:conf/cvpr/ChenH21}                 & \multicolumn{1}{c|}{13.07}    & 94.09   \\
\textbf{SupCon}~\cite{DBLP:conf/nips/KhoslaTWSTIMLK20}                  & \multicolumn{1}{c|}{\textbf{10.92}}    & \textbf{95.18}   \\ \hline
\end{tabular}
}
\caption{Comparisons of the different contrastive methods.}
\label{tab:contrasitve}
	\end{minipage}
	\hfill
	\begin{minipage}{0.40\linewidth}
	\centering
\scalebox{0.6}{
\begin{tabular}{c|cc}
\hline
\multirow{2}{*}{Loss Design} & \multicolumn{2}{c}{I\&C\&M to O}        \\ \cline{2-3} 
                             & \multicolumn{1}{c|}{HTER(\%)} & AUC(\%) \\ \hline
None                         & \multicolumn{1}{c|}{13.14}    & 93.09   \\ 
\textbf{Domain Entropy}             & \multicolumn{1}{c|}{\textbf{10.92}}    & \textbf{95.18}   \\
MMD Loss                     & \multicolumn{1}{c|}{12.26}    & 93.17   \\
Cosine Distance              & \multicolumn{1}{c|}{13.05}    & 93.95   \\ \hline
\end{tabular}
}
\caption{Comparisons of different domain generalized methods.}
\label{tab:domain_loss}
	    \end{minipage}
\vspace{-20pt}
\end{table}

\subsection{Abalation Studies}
In this subsection, we conduct ablation experiments on our designed modules to investigate the effectiveness of each component. Since the protocol I\&C\&M to O is the most challenging, we report results based on this protocol.

\noindent \textbf{Ablation of Components.} In Table \ref{table:ab_module}, we ablation each component of EnfoMax, i.e., supervised contrastive and domain classification modules. When we only add the supervised contrastive or domain classification module, the model's performance slightly improves in each protocol. However, when the domain classification and supervised contrastive modules are added, the model's performance is significantly improved. This result verifies that the proposed modules can benefit the encoder in learning more domain-invariant features of live samples. We also notice that when we only use reconstruct module, the model's performance is unsatisfactory because, at this point, the model only fits the source domain data and does not have good generalization capability.

\noindent \textbf{Effectiveness of Contrastive Module.} In order to investigate the impact of the supervised contrastive module, a comparison is made with other contrastive methods presented in Table \ref{tab:contrasitve}. We keep the domain classification module and only change different contrastive methods. The result in Table \ref{tab:contrasitve} indicates that our supervised contrastive method performs best. In addition, other contrastive methods~\cite{DBLP:conf/cvpr/ChenH21,DBLP:conf/icml/ChenK0H20} need another encoder as the dictionary, which introduces additional training costs. We argue that the supervised contrastive module is more suitable for FAS tasks since the other contrastive methods always use different data augmentation of the same image to form a positive sample pair. All other images are treated as negative samples. However, in the FAS task, we want to find common features of the live samples, and the above comparison methods do not serve to pull in all the live samples. Therefore, we use the supervised contrastive module in our EnfoMax framework.   


\noindent \textbf{Effectiveness of Domain Classification Module.} Domain classification module maximizes the domain entropy of living samples from different source domains. We use this module to remove the domain-special information contained in the feature. However, except for domain entropy, many designs of this similar function also exist. We compare them in Table \ref{tab:domain_loss}. As shown in Table \ref{tab:domain_loss}, the domain entropy norm method performs best. In addition, we found that no matter which constraint loss is added, the model's performance can be improved. This shows that removing the domain-special information in the feature can improve the domain generalized performance of the model when the target domain is unavailable.

\begin{table}[]
\centering
\setlength{\tabcolsep}{3mm}
\scalebox{0.8}{
\begin{tabular}{c|cc}
\hline
\multirow{2}{*}{Method} & \multicolumn{2}{c}{I\&C\&M to O}             \\ \cline{2-3} 
                        & \multicolumn{1}{c|}{HTER(\%)}   & AUC(\%)    \\ \hline
SSDG-R~\cite{jia2020single}                  & \multicolumn{1}{c|}{16.76$\pm$1.33} & 91.72$\pm$1.26 \\
SSAN-R ~\cite{Wang_2022_CVPRSSAN}                 & \multicolumn{1}{c|}{25.72$\pm$3.74} & 79.37$\pm$4.64 \\
PatchNet  ~\cite{Wang_2022_CVPR}              & \multicolumn{1}{c|}{23.49$\pm$1.90} & 84.62$\pm$1.92 \\
DiVT-T  ~\cite{DBLP:conf/wacv/LiaoCLYHC23}              & \multicolumn{1}{c|}{16.63$\pm$1.36} & 90.58$\pm$1.30 \\
\textbf{EnfoMax}                 & \multicolumn{1}{c|}{\textbf{13.18$\pm$1.29}} & \textbf{93.25$\pm$1.20} \\ \hline
\end{tabular}
}
\caption{Comparisons of average performance of the model after convergence.}
\label{tab:conv}
\vspace{-20pt}
\end{table}

\noindent \textbf{Convergence stability.}
Since most of the previous methods report the best results when the model is not fully converged in the source domain. The results thus obtained have a certain degree of chance since most previous methods are heuristic and lack relevant theoretical analysis. Therefore, we compare different methods' performance from the last ten epochs upon the model converging in the source domain. The result is shown in \ref{tab:conv}. We notice that when the model converges in the source domain, EnfoMax performs best since our method has the guarantee of the theoretical lower bound. Meanwhile, the adversarial learning-based method SSAN~\cite{Wang_2022_CVPRSSAN} fluctuates greatly compared to other methods, illustrating the instability of adversarial training. 

\noindent \textbf{Different Backbones.}
Since our method can be used in different backbones, we evaluate our EnfoMax method in various ViT backbones, which also be used in domain generalized FAS tasks~\cite{wang2022transfas,DBLP:conf/wacv/LiaoCLYHC23,DBLP:journals/tifs/WangWDG22}. In addition, we compare the performance of other domain generalized FAS methods that rely on the same ViT backbones as our approach, as presented in Table \ref{tab:bacobone}. Our EnfoMax method achieves the best performance, which shows that our EnfoMax method can be applied to various ViT backbones.

\noindent \textbf{Other proxy task for estimating $I^{s}(T;Y)$.}
The reconstruct task serves as the proxy task in our EnfoMax for the purpose of maximizing the lower bound of Equation \ref{low1}. Notably, another proxy task exists for lowering the $I^{s}(T;Y)$~\cite{DBLP:journals/corr/abs-1907-02893}. , which employs the representation feature $T$ to minimize the empirical risk on the source domain. We present a comparison of these two proxy tasks in Table \ref{tab:diffproxy}. The result shows that reconstruct task as the proxy task has better performance.

\begin{table}[t]
\centering
\setlength{\tabcolsep}{3mm}
\scalebox{0.9}{
\begin{tabular}{c|cc}
\hline
\multirow{2}{*}{Proxy Task} & \multicolumn{2}{c}{I\&C\&M to O}        \\ \cline{2-3} 
                            & \multicolumn{1}{c|}{HTER(\%)} & AUC(\%) \\ \hline
\textbf{Reconstruct}                 & \multicolumn{1}{c|}{\textbf{10.92}}    & \textbf{95.18}   \\
Empirical risk              & \multicolumn{1}{c|}{11.53}    & 94.12   \\ \hline
\end{tabular}
}
\caption{Comparisons of different proxy tasks.}
\label{tab:diffproxy}
\end{table}

\begin{table}[]
\centering
\setlength{\tabcolsep}{3mm}
\scalebox{0.8}{
\begin{tabular}{c|cc|cc}
\hline
\multirow{2}{*}{Method} & \multicolumn{2}{c|}{ViT-Small}                        & \multicolumn{2}{c}{ViT-Base}                        \\ \cline{2-5} 
                        & \multicolumn{1}{c|}{HTER(\%)}       & AUC(\%)        & \multicolumn{1}{c|}{HTER(\%)}       & AUC(\%)        \\ \hline
ViT${^\dag}$ ~\cite{dosovitskiy2020image}                    & \multicolumn{1}{c|}{20.85}          & 84.92          & \multicolumn{1}{c|}{26.65}          & 77.91          \\
MAE~\cite{MaskedAutoencoders2021}                     & \multicolumn{1}{c|}{14.72}          & 92.67          & \multicolumn{1}{c|}{18.47}          & 88.28          \\
DiVT~\cite{DBLP:conf/wacv/LiaoCLYHC23}                    & \multicolumn{1}{c|}{14.27}          & 93.62          & \multicolumn{1}{c|}{21.37}          & 86.24          \\
\textbf{EnfoMax}        & \multicolumn{1}{c|}{\textbf{11.21}} & \textbf{94.74} & \multicolumn{1}{c|}{\textbf{14.44}} & \textbf{91.96} \\ \hline
\end{tabular}
}
\caption{Comparisons of different backbone based on protocol I\&C\&M to O. The ${\dag}$ means the model is pre-trained using the ImageNet dataset. }
\label{tab:bacobone}
\end{table}

\begin{figure}[ht]
\centering
\includegraphics[width=0.45\textwidth]{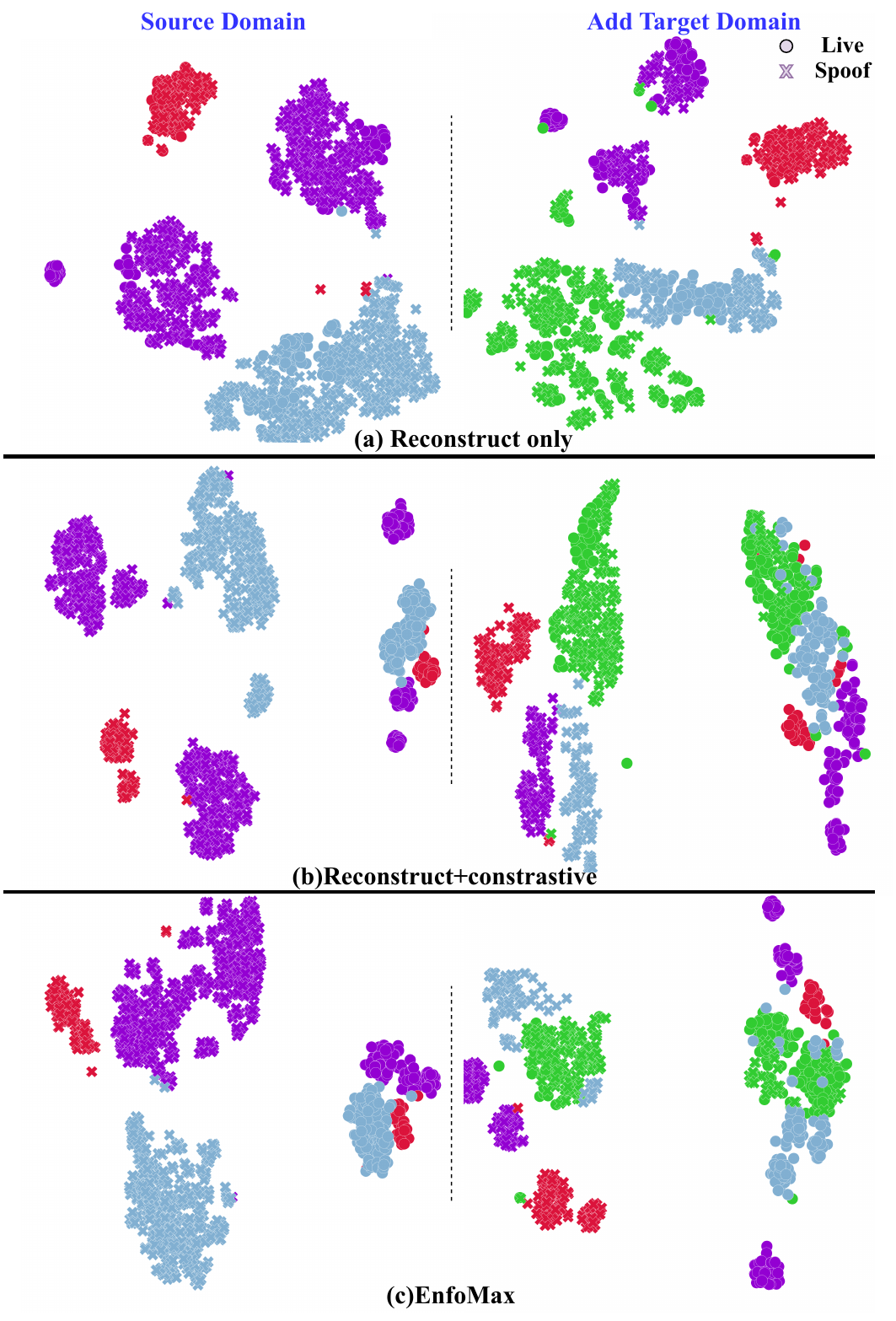} 
\caption{Visualizations of the feature distribution of di, EnfoMax and EnfoMax without domain classification module in cross-domain protocol I\&C\&M to O. The circle represents the live sample, and the cross represents the spoof sample. We use the pre-trained model in protocol I\&C\&M to O. The left column is the feature of the source domain dataset, and the right column is the feature of all domain datasets. Purple represent dataset \textcolor[RGB]{148,0,211}{Replay-Attack(I)}, red represent dataset \textcolor[RGB]{220,20,60}{MSU-MFSD(M)}, blue represent dataset \textcolor[RGB]{130,176,210}{CASIA-MFSD(C)} and green represent target dataset \textcolor[RGB]{50,205,50}{OULU-NPU(O)}.}
\label{fig:tsne}
\end{figure}

\subsection{Visualization and Analysis}
\textbf{t-SNE.} To illustrate the effectiveness of our methods, we visualize the feature of the EnfoMax trained model using t-SNE in Figure \ref{fig:tsne}. We find apparent gaps in face images from different domains when we only use the reconstruction task to train the model. Since in that case, we did not give label information, so live and spoof samples cannot be distinguished. Meanwhile, there is also a certain domain gap between the target domain and the source domain. The live and spoof samples are well distinguished in the source domain in EnfoMax. Due to the addition of the domain classification module, the distance between the live samples from different domains is very small. After adding the data of the target domain, most of the target live samples can be correctly classified and mixed well with the source domain data. When we removed the domain classification module of EnfoMax, Even though the live and spoof samples are still well distinguished in the source domain, there is a certain distance between live samples from different domains. Meanwhile, the target and source domain data are not well aggregated. The above phenomenon further illustrates the importance of the domain classification module.

\begin{figure}[t]
\centering
\includegraphics[width=0.3\textwidth]{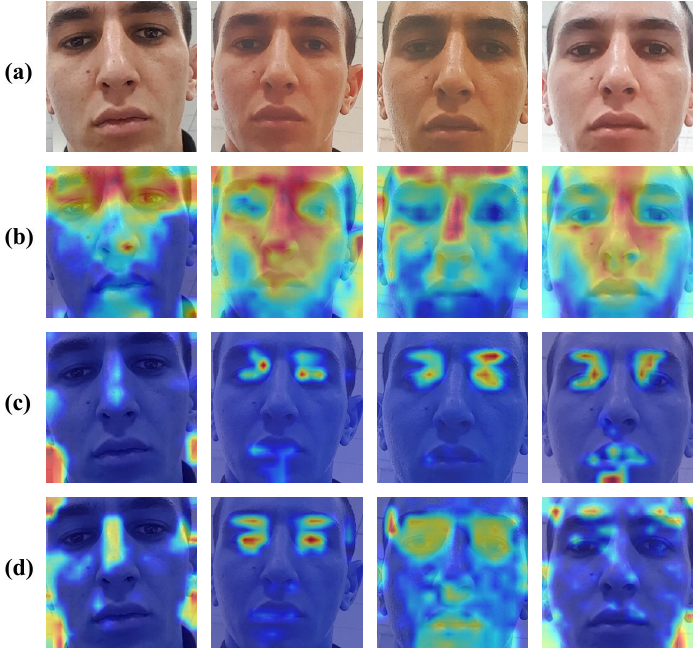} 
\caption{Attention visualization of different methods. (a): Original images. (b): EnfoMax method. (c): Reconstruct only method. (d): Vanilla ViT. Compared with other methods, the EnfoMax method focuses more on global features. }
\label{fig:cam}
\end{figure}

\noindent \textbf{Attention Maps.}
To identify the regions of the face that the model focuses on, we employ grad-cam~\cite{zhou2016learning} to produce the activation maps on the original images. We show the result in Figure \ref{fig:cam}. We find that reconstruct only and vanilla ViT method pay attention to the local feature of the input faces. On the contrary, EnfoMax pays attention to the global feature of the input face image. This means that the EnfoMax model can more effectively utilize the global features of the face image, thereby obtaining better generalization.

\section{Conclusion}
In this paper, we utilize mutual information and domain entropy to analyze the cross-domain face anti-spoofing (FAS) task, which provides theoretical guarantees and optimization objectives for domain generalized FAS tasks. We aim to maximize the mutual information between the feature and label in the unseen target domain for domain generalized FAS tasks. Since the domain gap always exists and the target information is unavailable, we use the source domain's mutual information and domain entropy to give a lower bound of the mutual information between the feature and label in the unseen target domain. To optimize the above lower bound, we present a novel learning framework called EnfoMax. Our method uses domain entropy to remove the domain-special features without adversarial learning, which has better stability compared to previous work. Experimental results show a significant performance improvement of the EnfoMax method in domain generalized FAS tasks. Furthermore, our information-theoretic analysis is independent of the proxy tasks employed, enabling the design of additional proxy tasks to enhance performance further.

\appendix

{\small
\bibliographystyle{ieee_fullname}
\bibliography{egbib}
}

\end{document}